# Deep-Learning Inversion of Seismic Data

Shucai Li, Bin Liu, Yuxiao Ren, Yangkang Chen, Senlin Yang, Yunhai Wang, *Member, IEEE*, and Peng Jiang, *Member, IEEE*

*Abstract*— We propose a new method to tackle the mapping challenge from time-series data to spatial image in the field of seismic exploration, i.e., reconstructing the velocity model directly from seismic data by deep neural networks (DNNs). The conventional way of addressing this ill-posed inversion problem is through iterative algorithms, which suffer from poor nonlinear mapping and strong nonuniqueness. Other attempts may either import human intervention errors or underuse seismic data. The challenge for DNNs mainly lies in the weak spatial correspondence, the uncertain reflection–reception relationship between seismic data and velocity model, as well as the time-varying property of seismic data. To tackle these challenges, we propose end-to-end seismic inversion networks (SeisInvNets) with novel components to make the best use of all seismic data. Specifically, we start with every seismic trace and enhance it with its neighborhood information, its observation setup, and the global context of its corresponding seismic profile. From the enhanced seismic traces, the spatially aligned feature maps can be learned and further concatenated to reconstruct a velocity model. In general, we let every seismic trace contribute to the reconstruction of the whole velocity model by finding spatial correspondence. The proposed SeisInvNet consistently produces improvements over the baselines and achieves promising performance on our synthesized and proposed SeisInv data set according to various evaluation metrics. The inversion results are more consistent with the target from the aspects of velocity values, subsurface structures, and geological interfaces. Moreover, the mechanism and the generalization of the proposed method are discussed and verified. Nevertheless, the generalization of deep-learning-based inversion methods on real data is still challenging and considering physics may be one potential solution.

## I. Introduction

SEISMIC exploration is often used to map the structure of subsurface formations based on the propagation of

Manuscript received March 26, 2019; revised July 29, 2019 and October 30, 2019; accepted November 4, 2019. Date of publication December 11, 2019; date of current version February 26, 2020. This work was supported in part by the National Natural Science Foundation of China under Grant 51739007, Grant 61702301, and Grant 51809155, in part by the Royal Academy of Engineering under the U.K.–China Industry Academia Partnership Program Scheme (U.K.-CIAPP\314), in part by the Key Research and Development Plan of Shandong Province under Grant 2016ZDJS02A01, and in part by the Fundamental Research Funds of Shandong University. *(Corresponding author: Shucai Li.)*

S. Li, B. Liu, Y. Ren, S. Yang, and P. Jiang are with the School of Qilu Transportation, Shandong University, Jinan 250100, China, and also with the Geotechnical and Structural Engineering Research Center, Shandong University, Jinan 250100, China (e-mail: lishucai@sdu.edu.cn; liubin0635@163.com; ryxchina@gmail.com; yangsenlin@mail.sdu.edu.cn; sdujump@gmail.com).
Y. Chen is with the School of Earth Sciences, Zhejiang University, Hangzhou 310027, China (e-mail: yangkang.chen@zju.edu.cn).
Y. Wang is with the School of Computer Science and Technology, Shandong University, Qingdao 266237, China (e-mail: cloudseawang@gmail.com).

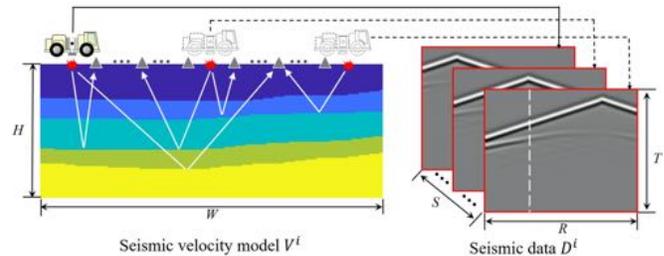

Fig. 1. Illustration of seismic exploration, which maps seismic velocity model (spatial image) to time-series seismic data. Seismic data are composed of seismic profiles indicated by a red border. Each seismic profile corresponds to seismic data recorded by all receivers from a seismic source. The white dashed line on the seismic profile indicates a seismic trace recorded by a single receiver.

seismic wave on the Earth. It can estimate the physical properties of the Earth's subsurface mainly from reflected or refracted seismic wave. Since seismic exploration is capable of detecting target features from a large to small scale, it plays an important role in the delineation of near-surface geology for engineering purposes, hydrocarbon exploration, as well as the Earth's crustal structure investigation. Usually, artificial sources of energy are required and a series of receivers are placed on the surface to record seismic waves (as shown in Fig. 1). One major outcome of processing the recorded data is the reconstruction of the subsurface velocity model, namely seismic velocity inversion, which has a substantial impact on the accuracy of locating and imaging target bodies. Recently, in addition to stochastic inversion strategies [1]–[3], by using full-waveform information of seismic data, full-waveform inversion (FWI) is now one of the most appealing methods to reconstruct the velocity model with high accuracy and resolution [4]–[7].

FWI was first proposed in the early 1980s. It reconstructs the velocity model by iteratively minimizing the difference between seismic data and synthetic data in a least-squares sense [8]–[12]. Conventional FWI uses gradient-based solvers to update the model parameters, and the gradient is normally calculated through backward wavefield propagation of data residuals based on adjoint-state methods [13]–[15]. However, seismic velocity estimation from observed signals is a highly nonlinear process, so that the conventional iterative algorithm usually requires a good starting model to avoid local minimum. Moreover, FWI also faces severe nonuniqueness because of inadequate observation or observation data contaminated with noise. Faced with these nonlinearity and nonuniqueness issues in conventional FWI, geophysicists have proposed many improvements, such as the multiscale strategy [16]–[19], processing of seismic data in other domains [20]–[22], and so on.

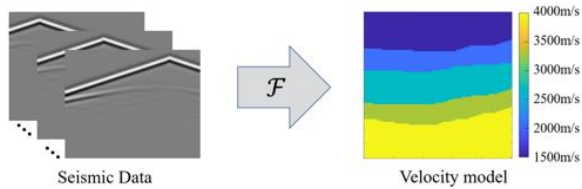

Fig. 2. Task definition. DNN-based seismic inversion is a method to learn the function $\mathcal{F}$ that maps seismic data to its corresponding velocity model.

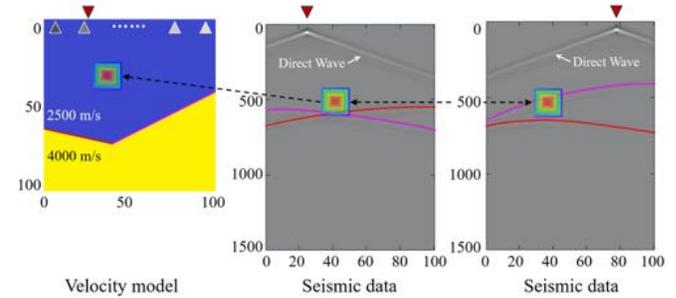

Fig. 3. Velocity model and its corresponding seismic profiles. (Left) Velocity model with one downward and upward interface. (Middle) Profile of seismic data based on one shot in location 25. (Right) Profile of seismic data based on one shot in location 75. The red and purple curves on the seismic profiles indicate signals from the downward and upward parts of the interface, respectively. Note that the squares with jet color indicate patches that all lie on the same relative positions. The patch on the left seismic profile contains reflected signals from both upward and downward parts of interface, whereas the patch on the right seismic profile contains only signals from the upward part. However, the patch on the velocity model does not have any information about the interface. That is, either spatial correspondence between the velocity model and the seismic profiles or between different seismic profiles is not aligned.

Recently, deep neural networks (DNNs) have demonstrated remarkable ability to approximate nonlinear mapping function between various data domains [23], such as images and label maps [24], [25], images and text [26], [27], and different types of images [28], [29], especially for inverse problems, such as model/image reconstruction [30], [31], image super-resolution [32]–[34], real-world photosynthesis [35], [36], and so on. These state-of-the-art development brings new perspectives for seismic inversion and velocity model reconstruction. DNN-based seismic inversion is to learn the mapping function $\mathcal{F}$ from seismic data to velocity model, as illustrated in Fig. 2. So far, some work has already made progress on this task. Moseley *et al.* [37] achieved 1-D velocity model inversion by WaveNet [38] after depth-to-time conversion of the velocity profile. Araya-Polo *et al.* [39] used convolutional neural networks (CNNs) to reconstruct velocity model from a semblance cube calculated from raw data. These two approaches may introduce biases because of the human intervention in seismic data processing. Other than data processing, Wu *et al.* [40] proposed InversionNet to build the mapping from raw seismic data to the corresponding velocity model directly by Autoencoder architecture, which decodes the velocity model from an encoded embedding vector. Since data are extremely condensed in the embedding vector, the decoded velocity model may lose details.

Inspired by recent progress and to avoid potential problems of the work mentioned above, we further analyze the characteristics of the DNN-based seismic inversion task. In general, the characteristics mainly consist of three aspects. First, the spatial correspondence between raw time-series signals (seismic data) and seismic images (velocity model) is weak, especially for reflected seismic signals. As illustrated in Fig. 3, the position for which there exists a reflected wave on the seismic profile, the corresponding position on the velocity model may not contain any interface and vice versa. This weak spatial correspondence issue was also mentioned in [41]. Second, considering the complexity of subsurface structure in various velocity models, for reflected seismic signals received by one receiver from one source (i.e., seismic trace), the corresponding interfaces which cause the reflections in velocity model are uncertain. In this article, we use the term uncertain reflection–reception relationship to refer to this characteristic. Third, in raw seismic data, the recorded seismic wave gradually weakens as time goes by. That is, seismic data are of time-varying property, which makes seismic pattern hard to capture with fixed kernel. All these characteristics pose a great challenge for DNNs, especially for CNNs with spatial correspondence and weight-sharing properties.

To address these unique and intrinsic characteristics, in this article, we propose an end-to-end DNN with novel components called the seismic inversion network (SeisInvNet). The idea is that we first learn a feature map spatially aligned to the velocity model from each seismic trace and after building the spatial correspondence, the velocity model could be reconstructed from the learned feature maps effortlessly. Specifically, considering the uncertain reflection–reception relationship, we enforce each seismic trace to learn a feature map with information corresponding to the whole velocity model regardless of actual relationship between each seismic trace and velocity model. Then, from all the feature maps, we reconstruct the velocity model by CNNs. Thus, after training, the information in each feature map will spatially align to the velocity model. Since each seismic trace has its own sensitivity region, the feature map from different seismic traces will provide knowledge to reconstruct different parts of a velocity model. We learn spatially aligned feature maps by fully connected layers, which can handle the uncertainty and time-varying properties of seismic data. In this way, we solve the challenge of uncertain reflection–reception relationship, build the spatial correspondence, and meanwhile get rid of the time-varying problem.

Moreover, in seismic exploration, a seismic trace is often observed along with its neighborhood traces to identify the local structure, and all seismic traces are combined together to deduce the global context. Thus, before learning the feature map, we enhance each seismic trace with some auxiliary knowledge from several neighborhood traces and the whole profile. Furthermore, all the mentioned operations could be trained end to end. In general, we make the best use of all seismic data and let every seismic trace contribute to the reconstruction of the whole velocity model by finding spatial correspondence.

We carry out experiments on a self-proposed SeisInv data set. From comprehensive comparisons, our SeisInvNet has demonstrated superior performance against all the baseline models consistently. Our inversion results are more consistent with the ground truth from the aspects of velocity values, subsurface structures, and geological interfaces. Furthermore, we study the mechanism of SeisInvNet by visualization and statistical analysis.

Major contributions reside in the following aspects.
1) We make an in-depth analysis on the problem of DNN-based seismic inversion.
2) We design SeisInvNet with novel and efficient components to take full advantage of all the seismic data.
3) We demonstrate superior performance against all the baseline models on the proposed SeisInv data set.
4) We provide comprehensive mechanism studies.

## II. Related Works

Several works have tried seismic inversion problems by DNNs in various ways, which can be concluded as follows.

### A. Data Preprocessing-Based Seismic Inversion

To make raw seismic data and the corresponding velocity model spatially aligned, data preprocessing is always an option to consider, though at the risk of introducing human assumptions. Moseley et al. [37] made a single-receiver recorded signal spatially aligned to the model by the standard 1-D time-to-depth conversion before performing 1-D velocity model inversion by WaveNet [38]. Araya-Polo et al. [39] transferred raw seismic data into a velocity-related feature cube by the normal moveout correction of common midpoint gather data [42] before applying CNNs.

### B. Encoder–Decoder Network-Based Seismic Inversion

When images have specific fixed structures or patterns, they can be reconstructed without providing low-level information, such as spatial correspondence. In this case, even discarding spatial information and compressing original data to a 1-D embedding vector, we can still get desired output without losing much information. This process can be achieved by the encoder–decoder networks (Autoencoder for short). An Autoencoder has two components: one is the encoder which compresses input to the embedding vector and the other is the decoder which reconstructs the output from the embedding vector. This kind of Autoencoder has been widely utilized for tasks on the aligned face data set, such as face generation [43], [44]. According to their experiments, the embedding vector contains high-level information, such as expression, gender, and age. It also fits the human behaviors that we can draw the portrait only given some high-level semantic descriptions. However, when it comes to more complex images, low-level information will have critical importance. For example, medical image semantic segmentation methods, such as U-Net [31], [45], concatenate low-level features on the output of deep layers by shortcuts to compensate details of the results, but they only work on data where input and output are spatially aligned.

For seismic inversion problem where input and output are not spatially aligned, encoder–decoder networks can still be applied to extract the embedding vector. This way was adopted by Wu et al. [40] to propose InversionNet, which works well on the velocity model with relatively fixed patterns (horizontal interfaces and dipping faults) according to their article and can successfully handle complex structures after introducing CRF as postprocessing in [46]. However, when velocity models contain a large number of small structures, the embedding vector may fail to preserve all the details. As Fig. 7 demonstrates, Autoencoder inverted results sometimes lose details in our implementation.

### C. Other Related Work

Multilayer perceptron (MLP) neural networks with fully connected layers are the most direct way to construct a mapping between data of weak spatial correspondence. The output of each fully connected layer will depend on all the input values, and thus, information carried by seismic data could be best captured. However, in this way, the computation and space complexity is proportional to the square of data dimension. Dahlke et al. [41] developed a probabilistic model to indicate the existence of faults in the 2-D velocity model using MLPs, while Araya-Polo et al. [47] extended this method to the 3-D case. To reduce computational complexity, they have to first convert the model to the low-dimensional pixel or voxel grid, so that the faults could only be coarsely identified. In addition to reflected seismic waves, transmitted waves can also be used to reconstruct velocity models. For example, based on the prestack multishot seismic traces of transmitted wave, Wang et al. [48] achieved satisfactory inversion results using fully convolutional networks.

## III. Methodology and Implementation

### A. Methodology

In order to better define the problem, let us first introduce some notation. We have $N$ velocity models $V^i, i \in N$ and the corresponding $N$ seismic data $D^i, i \in N$. Each velocity model $V^i$ is of size $[H, W]$, whereas each seismic data $D^i$ is of size $[S, T, R]$. Here, $H$ and $W$ denote the height and width of the velocity model, whereas $S$, $R$, and $T$ denote the number of seismic sources, receivers, and time steps, respectively. Each seismic data $D^i$ can be treated as $S \times R$ single seismic trace $D^i_{s,r}$ of dimension $T$, while all the single seismic traces from the same source form the seismic profile $D^i_{s,:}$ of dimension $[T, R]$. Visual illustration of these notation is shown in Fig. 1.

In this article, we intend to learn the mapping $\mathcal{F}$ from seismic data $D^i$ to velocity model $V^i$ directly, namely seismic inversion, by DNNs with parameters $\theta$

$$\mathcal{F}(D^i, \theta) \to V^i. \quad (1)$$

In general, for an image-to-image mapping, CNNs are preferred. However, in our case, directly using CNNs may not be the optimal choice. There are two unique characteristics of the mapping between seismic data and velocity model.
1) Weak spatial correspondence. The interface in velocity model and the corresponding pattern in seismic data have weak spatial correspondence.
2) Uncertain reflection–reception relationship between seismic data and velocity model.

For a seismic trace, the corresponding interfaces that cause the reflected signals are uncertain in different velocity models. These characteristics will be problematic for the spatial correspondence property of CNNs. In addition, there is another potential problem that seismic waves weaken gradually as time goes by, which may pose another challenge for the weight-sharing property of CNNs. As stated in Section II, one possible way of using CNNs regardless of the spatial correspondence and weight sharing issue is by encoder–decoder networks, which condense seismic data into a 1-D embedding vector and abandon its spatial information. As demonstrated in Fig. 7, this way sometimes leads to inaccurate inversion of velocity models.

Instead, in this article, we intend to take full advantage of seismic data without much loss of information by DNNs. Consequently, we have the following methodology. In general, we first learn that the feature map $f_{s,r}^i$ contains information spatially aligned to the velocity model $V^i$ from each seismic trace $D_{s,r}^i$, and then we regress the velocity model $V^i$ from all the spatially aligned feature maps $[f_{s,r}^i : s \in S, r \in R]$. In practice, considering the uncertain relationship between seismic traces and interfaces in velocity model, regardless of actual relationship, we choose to let each seismic trace learn the feature map corresponding to the whole velocity model. After training, the information in each feature map will spatially align to the velocity model. As each seismic trace has its own sensitivity region, the feature map from different seismic traces will provide knowledge to reconstruct different parts of the velocity model. However, this mapping is ambiguous since different seismic observation setup and subsurface geology conditions may result in the same seismic trace records, i.e., $D_{\hat{s},\hat{r}}^i = D_{\bar{s},\bar{r}}^i$, $\hat{s} \neq \bar{s}$, and $\hat{r} \neq \bar{r}$. To reduce the uncertainty and enrich the knowledge, for each seismic trace data $D_{s,r}^i$ recorded by a single receiver $r$ and single shot $s$, we enhance the trace data by encoding its neighborhood information $\mathcal{N}(D_{s,:}^i)_r$, its observation setup $\mathcal{S}(D_{s,r}^i)$, and global context of its $s$th profile $\mathcal{G}(D_{s,:}^i)$, where we replace $D_{s,r}^i$ with an embedding vector $\mathbf{E}_{s,r}^i$, which gives

$$\mathbf{E}_{s,r}^i = \left[\mathcal{N}(D_{s,:}^i)_r, \mathcal{S}(D_{s,r}^i), \mathcal{G}(D_{s,:}^i)\right]. \quad (2)$$

Compared with $D_{s,r}^i$, $\mathbf{E}_{s,r}^i$ provides much more rich knowledge to generate spatially aligned and unambiguous feature map $f_{s,r}^i$. Intuitively speaking, neighborhood information would help networks become aware of the pattern of local seismic data, such as the existence and direction of reflected wave; observation setup would make information more distinguishable and less ambiguous by telling how each seismic trace data are recorded; and global context would supplement global information, such as velocity distribution and the number of interfaces.

Given this information, it is feasible to generate spatially aligned feature $f_{s,r}^i$ from $\mathbf{E}_{s,r}^i$ that

$$\mathcal{F}_1(\mathbf{E}_{s,r}^i, \theta_1) = f_{s,r}^i \quad (3)$$

where $\mathcal{F}_1$ and $\theta_1$ are the feature generating functions and its parameters. Collecting all the feature maps, we have $\mathbf{F}^i = [f_{s,r}^i : s \in S, r \in R]$ with dimension $[S \times R, h, w]$.

Finally, we could regress velocity model $V^i$ from $\mathbf{F}^i$ with commonly used CNNs, since we have built the spatial correspondence between $\mathbf{F}^i$ and $V^i$ that

$$\mathcal{F}_2(\mathbf{F}^i, \theta_2) = V^i \quad (4)$$

where $\mathcal{F}_2$ and $\theta_2$ are the velocity model regressing function and its parameters.

In a nutshell, in our method, we build the mapping $\mathcal{F}$ by

$$\mathcal{F} : D^i \xrightarrow{\mathcal{N}, \mathcal{S}, \mathcal{G}} \mathbf{E}^i \xrightarrow{\mathcal{F}_1} \mathbf{F}^i \xrightarrow{\mathcal{F}_2} V^i. \quad (5)$$

In the following, we will describe how to implement $\mathcal{N}, \mathcal{G}, \mathcal{F}_1, \mathcal{F}_2$ by DNNs.

### B. Implementation

Fig. 4 illustrates our proposed SeisInvNet which has four components.

*1) Embedding Encoder:* The embedding encoder generates embedding vectors that contain neighborhood information, observation setup, and global context.

We extract neighborhood information by a shallow CNN ($\mathcal{N}$) on the seismic profile $D_{s,:}^i$. The output of $\mathcal{N}$ has the same dimension as the input, but each value in $\mathcal{N}(D_{s,:}^i)$ shares information across its neighborhood in $D_{s,:}^i$. Consequently, the column $\mathcal{N}(D_{s,:}^i)_r$ (of dimension $T$) contains the neighborhood information of seismic trace $D_{s,r}^i$. As for observation setup, $\mathcal{S}$ transforms the position of the receiver $r$ and source $s$ to the one-hot vector. Since $s \in [1, S]$ and $r \in [1, R]$, the observation setup $\mathcal{S}(D_{s,r}^i)$ is a vector of dimension $S + R$. Global context is a vector of dimension $C$ extracted from seismic profile $D_{s,:}^i$ by an encoder $\mathcal{G}$. The encoder $\mathcal{G}$ is also implemented by CNNs, which constantly compresses data by convolution operation until the spatial dimension vanished.

Finally, as (2), we collect $\mathcal{N}(D_{s,:}^i)_r$, $\mathcal{S}(D_{s,r}^i)$, and $\mathcal{G}(D_{s,:}^i)$ to form an embedding vector $\mathbf{E}_{s,r}^i$ of dimension $T+S+R+C$, which replaces the original $T$-dimensional seismic trace $D_{s,r}^i$. Sometimes, we will refer $\mathbf{E}_{s,r}^i$ as enhanced seismic trace. It is worth to note that all CNNs based on $\mathcal{N}$ and $\mathcal{G}$ are weight sharing over all the applied data.

*2) Spatially Aligned Feature Generator:* Given the $S \times R$ embedding vectors $\mathbf{E}^i$, our generator first further condense them to vectors of size $h \times w$ using MLPs ($\mathcal{F}_1$) with several fully connected layers (including activation and norm operations), then reshape each vector to a feature map $f_{s,r}^i$ of size $[h, w]$. To be specific, instead of using raw seismic data (of size $[S, R, T]$) as a 3-D input, we treat each enhanced trace (of size $[T + S + R + C]$) as the actual input and let the parameters of the generator fixed and shared on all $[S \times R]$ of them. Thus, the mapping from each embedding vector to the feature map becomes acceptable for the generator with fully connected layers.

We design the generator to output feature map with the same dimension ratio as the velocity model. From these feature maps, the velocity model will be reconstructed directly. Thus, after training, every part of the feature map will impart knowledge to reconstruct the corresponding part of the velocity model, so feature map is spatially aligned to the velocity model. In addition, because each embedding vector has its own sensitivity region, a feature map $f_{s,r}^i$ learned from a different embedding vector would contribute to a different part of the velocity model. We show feature maps in Section IV-F to verify the above statements.

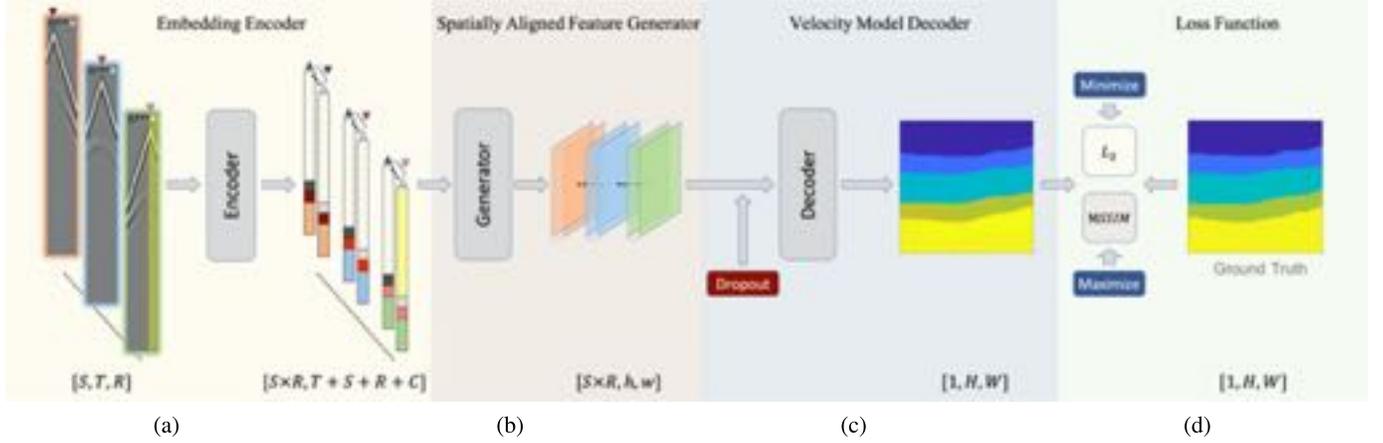

Fig. 4. Visualization of SeisInvNet framework. Given the seismic data (▼ to ▼ indicate data by different sources, while ▲ to ▲ indicate data recorded by different receivers. To save space, data are not visualized in its original scale.) (a) *Embedding encoder* replaces each original seismic trace $D_{s,r}^i$ by an embedding vector $\mathbf{E}_{s,r}^i$ which comprises neighborhood information $\mathcal{N}(D_{s,:}^i)_r$ (indicated by ☐ and one of the correspondences is indicated by the yellow region.), observation setup $\mathcal{S}(D_{s,r}^i)$ (indicated by squares, such as ■ and ■), and global context $\mathcal{G}(D_{s,:}^i)$ of the corresponding seismic profile (indicated by rectangles, such as ▮, ▮, and ▮). (b) *Spatially aligned feature generator* transforms each embedding vector to a feature map whose information is spatially aligned to the velocity model. (c) *Velocity model decoder* collects all the feature maps from which knowledge is decoded to regress the velocity model. (d) We optimize parameters of an encoder, a generator, and a decoder by minimizing $L_2$ and maximizing MSSIM metrics to make output more closed to the ground truth. Check Section III-B for details.

*3) Velocity Model Decoder:* A velocity model decoder collects all $S \times R$ feature maps $\mathbf{F}^i$ from which knowledge is decoded to regress the velocity model $V^i$ (of size $[1, H, W]$) by CNNs ($\mathcal{F}_2$) with several convolutional layers (including activation and norm operations). During training, the velocity model decoder randomly throws away several feature maps (dropout) to help the decoder reconstruct the velocity model from all the possible information, other than only depended on certain feature maps. The dropout operation improves robustness and generalization ability of the velocity model decoder.

*4) Loss Function:* How the reconstructed velocity models appear mainly depends on how the loss function is designed. For image regression and reconstruction problems, $L_1$ or $L_2$ norm is the most commonly used metric to define the loss function. However, these metrics treat each position of the image individually, which makes it difficult for networks to capture local structures and details, such as edges and corners. In the experiments of Wang *et al.* [49], with the same $L_2$ score, processed images show dramatic difference, which means $L_2$ is inconsistent with human perception.

Local structures and details are important factors to be taken into consideration while reconstructing seismic velocity models [50]–[52]. To measure these factors, the structural similarity index (SSIM) proposed by Wang *et al.* [49] is the most commonly used metric which computes the statistical difference between two corresponding windows, that is. predict and target. SSIM is defined as

$$\text{SSIM}(x, y) = \frac{(2\mu_x\mu_y + c_1)(2\sigma_{xy} + c_2)}{(\mu_x^2 + \mu_y^2 + c_1)(\sigma_x^2 + \sigma_y^2 + c_2)} \quad (6)$$

where $x$ and $y$ are the two corresponding windows, while $c_1$ and $c_2$ are the variables to stabilize the division. SSIM score ranges from 0 to 1, and it reaches its maximum when information in two windows are the same.

Accordingly, apart from minimizing norm metric, in this article, we will simultaneously maximize SSIM metric. Specifically, we apply squared $L_2$ norm and multiscale structural similarity (MSSIM) [53] to compute the loss $\mathcal{L}_i$ for each data pair

$$\mathcal{L}_i(\bar{V}^i, V^i) = \sum_{k=0}^{H \times W} \left(\bar{V}_k^i - V_k^i\right)^2 - \sum_{r \in R} \sum_{k=0}^{H \times W} \lambda_r \text{SSIM}\left(\bar{V}_{x(k,r)}^i, V_{y(k,r)}^i\right) \quad (7)$$

where $\bar{V}^i$ and $V^i$ are the inversion result and ground truth, respectively, for the $i$th data pair, and $x_{(k,r)}$ and $y_{(k,r)}$ are the two corresponding windows centered in $k$ of size $r$. We carry out SSIM on total $R$ different scales, and for each scale $r$, we have a weight $\lambda_r$ to control its importance. For the gradient derivation of SSIM, please refer to [54].

## IV. EXPERIMENTS

### A. Data Set Preparation

In this section, we give a detailed description of how the data set is prepared. Our data set has 12 000 different velocity models and the corresponding synthetic seismic data pairs. Since this data set is mainly for inversion problems, we name our data set SeisInv data set. Consequently, synthetic seismic data are the input, whereas the velocity model is the ground truth.

In this article, velocity models are designed to have horizontally layered structures with several small ups and downs randomly distributed along the interfaces. In general, the velocity models in our data set can be divided into four subsets, namely types I–IV, respectively, according to the number of subsurface interfaces. In each category, 3000 different models are first designed by generating the geology interfaces and then filling the layers between two adjacent interfaces with a constant velocity value randomly selected from [1500, 4000]. According to the statistics of actual subsurface geological conditions, the deeper stratum tends to have a larger seismic velocity. Consequently, we let the velocity value monotonically increase with stratum and the velocity difference is over 300 m/s for the adjacent geological layers. The overall statistics, including mean, variance, and histogram of values are demonstrated in Fig. 5(a) and (b). In this article, we design velocity models to have up to four interfaces, i.e., two to five layers with different velocity values. In order to obtain a more recognizable pattern on observation data, the interfaces are kept away from the top ten grids and mainly distributed in the middle, as shown in Fig. 5(c).

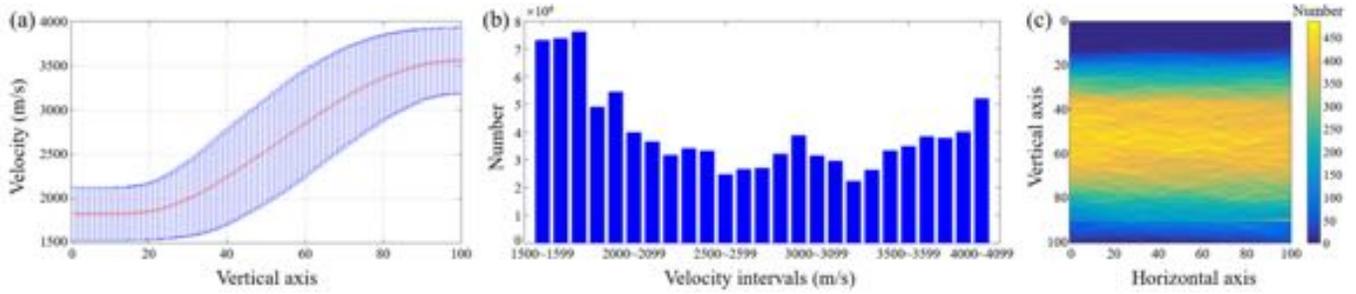

Fig. 5. Overall statistics on the training data set. (a) Velocity distribution with respect to the depth. Red stars: mean of velocities distributed in depth on the vertical axis. Blue intervals: the corresponding standard deviation. (b) Histogram of values in velocity models. Blue bars: the number of grids with a velocity value in the corresponding velocity interval. (c) Interface distribution on velocity models. The number of interfaces in each grid is visualized with the right-side colormap.

Seismic data are generated through numerical simulation of seismic wave propagation on velocity models. In seismic exploration, a series of seismic sources and receivers are set on the ground surface to excite and record seismic wave, respectively. Usually, the recorded signal will contain various types of waves, such as reflected waves, and diffracted waves. Normally, the primary reflected wave is the main signal used to extract information regarding subsurface structures. In the numerical generation of seismic data in this article, the velocity model is of grid size $100 \times 100$ with uniform spacing of 10 m on both directions, and 20 seismic sources are set on the surface uniformly with five-grid interval, while 100 receivers are set on every grid of the surface. Typically, acoustic wave equation [see (8)] is considered as the governing physics controlling the seismic wavefield changes in time

$$\frac{\partial^2 p}{\partial t^2} - v^2 \left( \frac{\partial^2 p}{\partial x^2} + \frac{\partial^2 p}{\partial y^2} \right) = 0 \qquad (8)$$

where $p$ denotes the pressure, i.e., the acoustic wavefield, $v$ denotes wave velocity of the media, $x$ and $y$ are the spatial coordinates, and $t$ is the time. Based on the pseudospectral method [55]–[57], seismic wave propagation in an arbitrary velocity model can be simulated accurately. The wavefield at every receiver point and every time step is saved as seismic data (two figures on the right-hand side in Fig. 3).

Note that, the first recorded wave, which has the biggest amplitude, is called the direct wave (as shown in Fig. 3). It represents waves that travel directly from seismic sources to the receivers, so it does not contain any information about subsurface structures. Below the direct wave there lie the primary reflected waves, which are indicated bys red and brown lines in Fig. 3. They have much smaller amplitude since part of the wave energy will go through the interface and continue to travel in the next medium. Even in the same medium, as time goes by, the amplitude of seismic record will gradually decrease due to dispersion and dissipation. This wavefield attenuation nature will pose a great challenge, especially for the CNNs that share uniform weights over the whole spatial dimensions.

### B. Experimental Settings

We randomly divide our SeisInv data set into three sets: 10 000 for training, 1000 for validation, and 1000 for test. To reduce the computation complexity and for the convenience of comparing with other methods, we only utilize the data of the front 1000 time steps and sample the data recorded by 32 receivers from the total 100 receivers uniformly. That is, the input data are from the same seismic acquisition setup. For each time step, the time sampling rate is 1 ms, and a study of how different time steps affect the final inversion results is carried out in Section IV-D3. Consequently, $D^i$ is of size [20, 1000, 32] and $V^i$ is of size [100, 100]. We apply MSSIM with five different scales from window size of 11 at the beginning, and the weight for each scale is set up identical with [53]. To optimize our SeisInvNet and baseline model, we use Adam optimizer [58] with batch size 12 and set initial learning rate as $5e$-5 following the "poly" learning rate policy. During training, we carry out 200 epochs in total and set the dropout rate to 0.2 in our velocity model decoder. The longest path in our SeisInvNet has 24 layers (each layer is a combination of convolution or fully connected operation, activation operation, and norm operation.). Our SeisInvNet has about 10M parameters and could be trained end to end without any preprocessing and postprocessing. Following the convention, we save the parameters that perform best on the validation set and carry out experiments on validation and test sets.

### C. Baseline Models

To verify the effectiveness of our SeisInvNet, we use fully convolutional networks as our baseline model. To be specific, we design the baseline model based on encoder–decoder networks as the ones used in [40] and implement it by our own code, since no official code is available yet. The baseline model has 21 layers, which are slightly fewer than our SeisInvNet. However, the amount of parameters in the baseline model is about 40M in our implementation that is three times larger than SeisInvNet because of frequently using convolution with large channels. To be fair, our SeisInvNet and baseline model are designed with the same basic network blocks and trained with the same hyperparameters and optimizer mentioned in Section IV-B. Please refer to the Supplementary Material for the detailed architecture of the SeisInvNet and baseline model.

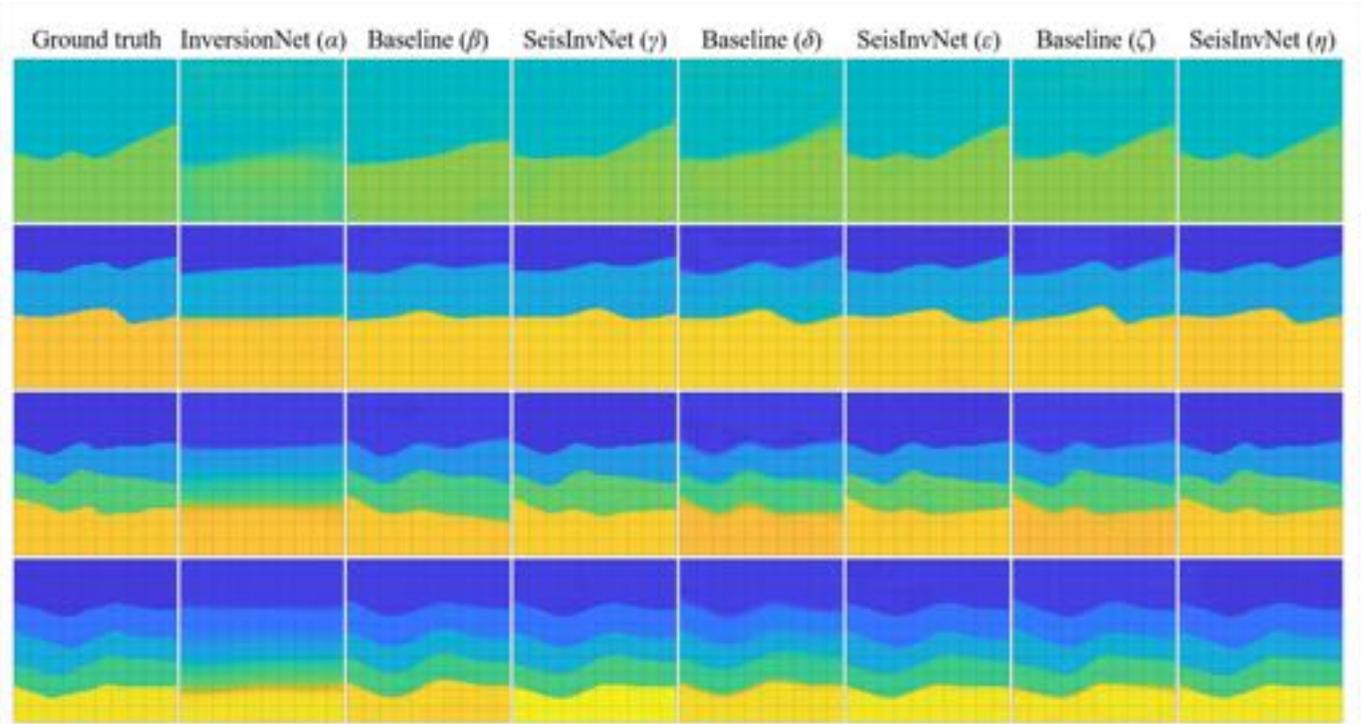

Fig. 6. Randomly selected examples from the test set. Each row from top to bottom exhibits examples from the velocity types I ∼ IV, respectively.

During comparison, we tested several variants of the SeisInvNet and baseline model by modifying loss function configuration. For each pair comparison of the SeisInvNet and baseline model, all the setups are identical. For example, in Table I, instead of comparing the SeisInvNet and baseline model with loss defined in (7), we also make comparisons with only $L_1$ loss and $L_2$ loss. It is worth to note that InversionNet [40] is a specific form of the baseline model with $L_1$ loss, initial learning rate as $5e$-4, and learning rate policy of decreasing ten times after every 15 training epochs. In our implementation, InversionNet may not converge to its optimal because of this dramatic learning rate decreasing. All the variants are trained, validated, and tested on the machine of a single NVIDIA TITAN Xp with similar average inference time of $0.013 \pm 0.002s$.

For convenience, we use a series of Greek letter to denote each variant of SeisInvNet and baseline model that ($\alpha$) for InversionNet, ($\beta$) for baseline with $L_1$ loss, ($\gamma$) for SeisInvNet with $L_1$ loss, ($\delta$) for baseline with $L_2$ loss, ($\epsilon$) for SeisInvNet with $L_2$ loss, ($\zeta$) for baseline with $L_2$ and MSSIM loss, and ($\eta$) for SeisInvNet with $L_2$ and MSSIM loss.

*D. Qualitative Comparison*

In this section, we will show some examples to visually demonstrate the inversion effects of our SeisInvNet variants, the baseline models, as well as the conventional FWI. In general, both DNN methods can successfully invert the velocity model, while SeisInvNet ($\eta$) presents more accurate value and detailed interface structures than the baseline model.

*1) Comparison With the Baseline Models:* From each type of velocity model in the test set, we randomly select several examples for comparison in Fig. 6. Apparently, among all the SeisInvNet variants and the baseline models, our SeisInvNet ($\eta$) can provide the best performance in most cases.

We further select several examples to visually compare SeisInvNet ($\eta$) and baseline ($\zeta$) more comprehensively in Fig. 7. The misfit degree of geological interface and velocity value are two major factors considered in this evaluation. As we can see, all four examples inverted by both SeisInvNet ($\eta$) and baseline ($\zeta$) present relatively uniform and accurate velocity distribution within every subsurface layer. Overall, observation indicates that the velocity model reconstructed by SeisInvNet ($\eta$) is closer to the ground truth from the aspects of velocity value, subsurface structure, and geological interface. To further analyze the inversion effects, we take a closer look at the interfaces and compare the inversion results at the pixel level. Generally speaking, baseline ($\zeta$) tends to present blurred interfaces, while SeisInvNet ($\eta$) generates a more accurate description of the interfaces, especially when the interfaces have some small undulations. Taking the example on the third row as an example, our SeisInvNet ($\eta$) successfully reconstructs the small undulation (marked by red circle) on the second interface, while baseline ($\zeta$) totally missed this structure and presented a flat interface. In addition, the results of baseline ($\zeta$) in the third and fourth rows have a poor inversion effect with anomalous velocity area and blurry interface (indicated by the red region), respectively, while the results of SeisInvNet ($\eta$) are more matched with the ground truth. A possible reason is that the baseline ($\zeta$) removes the spatial dimension, while our SeisInvNet ($\eta$) preserves spatial information, as discussed in Section IV-F.

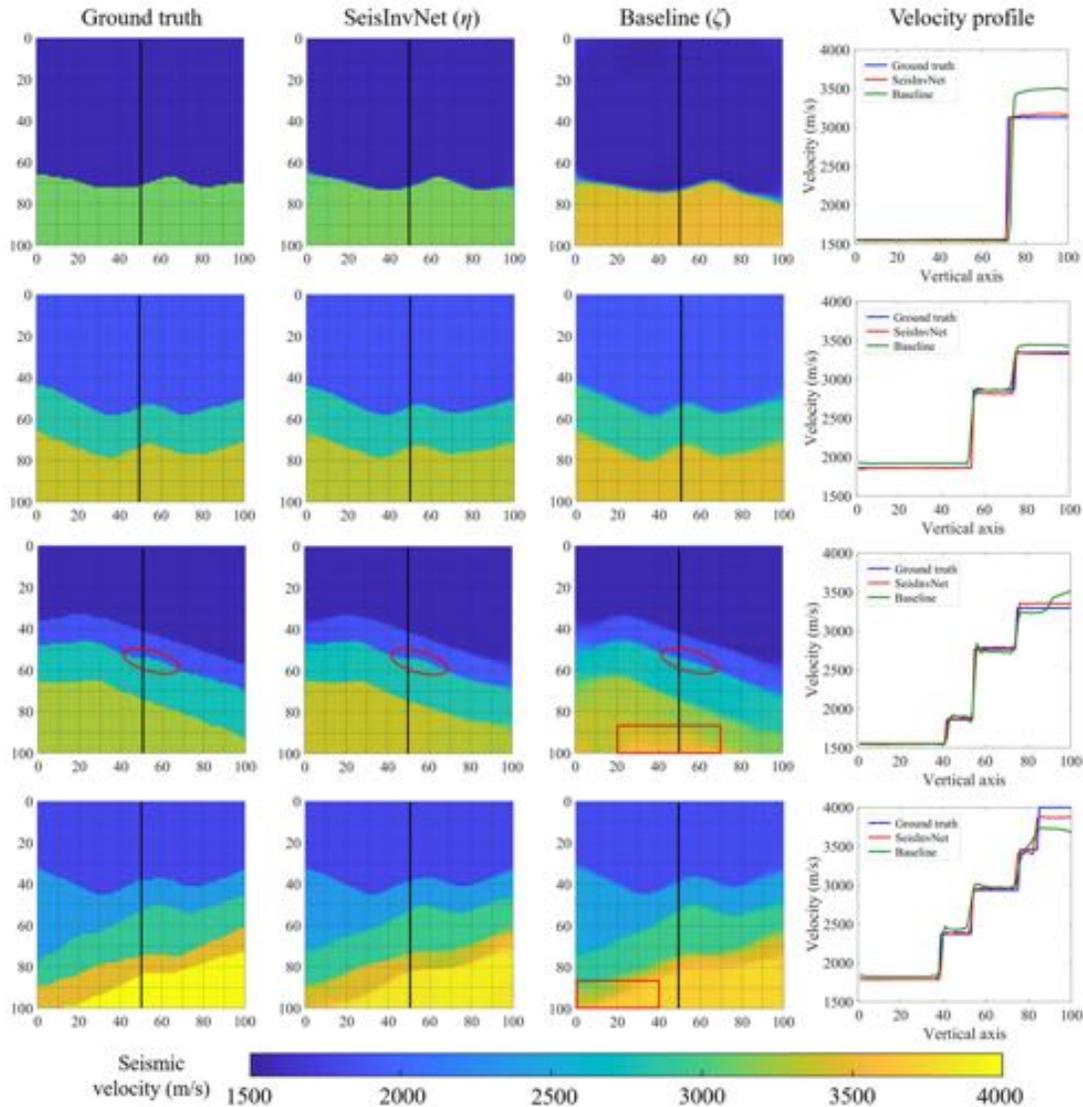

Fig. 7. Inversion results of our SeisInvNet (the second column) and baselines (the third column), as well as ground truth (the first column) from the validation and test sets. Vertical velocity profile at the black line is shown in the rightmost column for comparing the inverted velocity values. Each row from top to bottom exhibits examples from the velocity types I ∼ IV, respectively. The red ellipses on the third row circled a small undulation in the velocity model and the corresponding inversion results and the red rectangles on the third and fourth rows denote anomalous velocity areas. The comparison in this figure indicates that our SeisInvNet ($\eta$) delivers a better performance than the baseline model ($\zeta$) does.

From the velocity profiles at the vertical central axis shown in Fig. 7, we could find that our SeisInvNet ($\eta$) is better at the velocity recovery than the baseline model. In shallow layers, the reconstructed velocity model by SeisInvNet ($\eta$) seems identical with the ground truth, while baseline ($\zeta$) occasionally produces errors. As for the deeper layers, both methods show some discrepancy regarding the ground truth, while the error by SeisInvNet ($\eta$) is obviously smaller. This pattern is in line with the common sense of geophysical exploration that the deeper the subsurface structure is, the harder it is to estimate. Moreover, the velocity jump in the outputs of SeisInvNet ($\eta$) seems more "vertical" than that in baseline ($\zeta$), which means interfaces predicted by SeisInvNet ($\eta$) are sharper.

*2) Comparison With the Conventional FWI:* To demonstrate the advancement of our SeisInvNet over conventional methods, we further show several visual comparisons between the results of the FWI and SeisInvNet ($\eta$) in Fig. 8. The FWI results are calculated via a basic time-domain finite-difference code with a common conjugate gradient solver. As mentioned above, a good FWI result depends on a relatively good initial model while the deep learning way does not require this prerequisite. However, in most of the practical cases, the prior velocity distribution is hard to obtain and is usually blur and far from the ground truth. Thus, in this experiment, we use greatly smoothed velocity as initial models for FWI and the observation setup is the same as the one described in Section IV-B. From the inversion results shown in Fig. 8, we can see that our SeisInvNet($\eta$) has a better performance than the conventional FWI, especially in terms of the interface and the velocity distribution.

TABLE I
PERFORMANCE STATISTICS OF DIFFERENT METHODS ON TEST AND VALID SETS BY THE FIVE METRICS. FOR EACH SET AND METRIC,
THE TOP THREE RESULTS ARE HIGHLIGHTED IN RED, BLUE, AND GREEN, RESPECTIVELY. THE ↑ INDICATES THE LARGER
VALUE ACHIEVED, THE BETTER PERFORMANCE IS, WHILE ↓ INDICATES THE SMALLER, THE BETTER

| Dataset | Metric | InversionNet ($\alpha$) | $L_1$ Loss | | $L_2$ Loss | | $L_2$ & MSSIM | |
|---|---|---|---|---|---|---|---|---|
| | | | Baseline ($\beta$) | SeisInvNet ($\gamma$) | Baseline ($\delta$) | SeisInvNet ($\epsilon$) | Baseline ($\zeta$) | SeisInvNet ($\eta$) |
| Valid | MAE ↓ | 0.039207 | 0.028016 | **0.013614** | 0.028571 | **0.014404** | 0.027016 | **0.014782** |
| | MSE ↓ | 0.005172 | 0.003578 | **0.001726** | 0.002772 | **0.001315** | 0.002539 | **0.001254** |
| | SSIM ↑ | 0.823177 | 0.870728 | **0.932215** | 0.887269 | **0.941135** | 0.914004 | **0.953491** |
| | MSSIM ↑ | 0.879223 | 0.93565 | **0.974569** | 0.949324 | **0.979259** | 0.969629 | **0.986109** |
| | Soft $F_\beta$ ↑ | 0.382764 | 0.566493 | **0.735308** | 0.617445 | **0.741195** | 0.671889 | **0.762558** |
| Test | MAE ↓ | 0.039713 | 0.027841 | **0.013625** | 0.028504 | **0.01437** | 0.027404 | **0.014962** |
| | MSE ↓ | 0.005234 | 0.00341 | **0.001702** | 0.002675 | **0.001272** | 0.002466 | **0.001226** |
| | SSIM ↑ | 0.824075 | 0.870953 | **0.931366** | 0.888029 | **0.941919** | 0.914169 | **0.95338** |
| | MSSIM ↑ | 0.878726 | 0.936208 | **0.97468** | 0.948736 | **0.979546** | 0.969744 | **0.9861** |
| | Soft $F_\beta$ ↑ | 0.383811 | 0.56627 | **0.727225** | 0.610375 | **0.741908** | 0.668579 | **0.761035** |

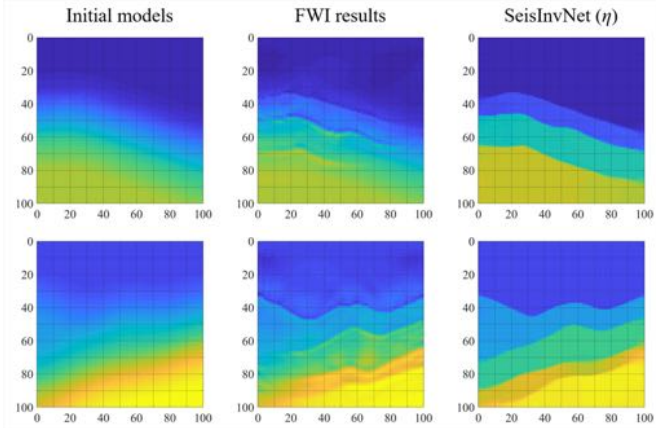

Fig. 8. Comparison between conventional FWI results with smooth velocity as initial model and the outcome of the SeisInvNet ($\eta$) (see the ground truth in Fig. 7).

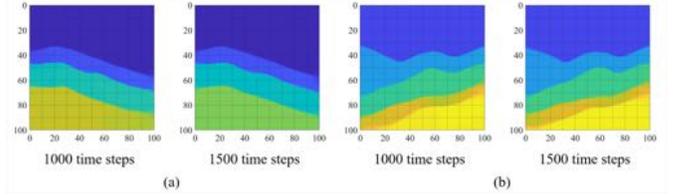

Fig. 9. SeisInvNet ($\eta$) outcomes with different recording times. (a) and (b) correspond to the type III and IV model in Fig. 7.

*3) Difficulty in the Recovery of the Last Layer:* The above visual comparisons show some difficulties in the recovery of the last layer. One possible reason is that the recording time is not long enough that the reflective information from the deeper parts of the model are not enough for velocity recovery. To verify this, we reprepare the data by extending the recording time to 1500 time steps and retrain our SeisInvNet from scratch. According to the evaluation results on the test set, the overall average $L_2$ loss decreases from 0.001226 to 0.001039, while the overall average MSSIM increases from 0.9861 to 0.9879, which shows small improvement and little visual quality promotion (see examples provided in Fig. 9). Therefore, we only use 1000 times steps in other experiments to reduce the number of parameters in networks and accelerate the speed of training and inference. Considering the above analysis, another possible reason would be that although most of the deep reflection waves are recorded, they are not strong enough to contribute in the velocity model recovery. How to extract the relatively weak signals from the deeper layers and how to better recover the deeper parts would be an essential topic worth exploring further in the future.

*E. Quantitative Comparison*

In this section, we quantitatively evaluate and compare the overall inversion effects of our SeisInvNet variants and baseline models via a series of metrics.

*1) Metrics:* Metrics we used are listed as follows.

*a) Mean Absolute Error (MAE) and Mean Square Error (MSE):* We quantify the misfit error of both inversion results based on the prevalently used MAE and MSE.

*b) SSIM and MSSIM:* We measure how well the local structures are fit by SSIM [49] and MSSIM [53].

*c) Soft $F_\beta$:* As stated previously, the quality of geological interfaces is one major factor considered when evaluating the inversion results. Geological interfaces appear as edges between two adjacent layers. Thus, we can first detect edges in the reconstructed velocity model and ground truth and then use the evaluation method for edge detection to measure their alignment. Namely, we measure the quality of geological interfaces by evaluating the accuracy of edges after applying the same edge detection method. And, we treat edges of $\bar{V}$ as predicted edges and edges of corresponding $V$ as ground-truth edges.

**Algorithm 1** Soft $F_\beta$

**Require:**
   (a) Reconstructed velocity model $\bar{V}$,
   (b) Ground-truth velocity model $V$,
   (c) Binarized edge detection method $E$.

**Steps:**
1: Compute binarized edges for both $\bar{V}$ and $V$ that $\bar{e} = E(\bar{V})$ and $e = E(V)$,
2: Smooth $\bar{e}$ by Gaussian kernel $\omega$ that $\hat{e} = \omega * \bar{e}$, and $*$ represents convolution filter,
3: *Soft Precision* $= |\hat{e} \odot e|/|\bar{e}|$, $\odot$ is a Hadamard product,
4: *Soft Recall* $= |\hat{e} \odot e|/|e|$, $\odot$ is a Hadamard product.

**Soft $F_\beta$:**
   Substitute *Precision* and *Recall* with *Soft Precision* and *Soft Recall* in Eq. 9, we get *Soft $F_\beta$*.

$F_\beta$ also called $F$-measure is a metric to measure the accuracy of classification, which is defined as

$$F_\beta = (1 + \beta^2) \cdot \frac{\text{Precision} \cdot \text{Recall}}{\beta^2 \cdot \text{Precision} + \text{Recall}} \quad (9)$$

and can also be used to measure the accuracy of edges, and it ranges from 0 to 1 with the rule of the larger the better that 1 for complete alignment and 0 reversely. However, this metric is too rigid that no matter how much predicted edges deviated from ground-truth edges, resulting in a total misalignment. This problem also exists in other metrics, such as MAE and *cross entropy*.

Consequently, we modify $F_\beta$ to a soft metric for edge detection that the more spatial closed the prediction to the target, more larger the $F_\beta$ will be (see definition in Algorithm 1). It is worth noting that since *Soft $F_\beta$* no longer follows its original formulation, the value range will change. In our experiments, the value range may exceed 1 but still follows the original rule. We set $\omega$ to 7 and $\beta$ to 1 in the following experiments.

*2) Results and Ablation Study:* The comparison results of all seven variants of SeisInvNet and baseline model by total five metrics on both validation (valid) and test sets are listed in Table I. Obviously, for each pair of SeisInvNet and baseline with the same loss function configuration, our SeisInvNet shows consistent superiority according to all the metrics. On the whole, our SeisInvNet ($\eta$) with $L_2$ and MSSIM loss achieved the best performance, since MAE and MSE, SSIM and MMSIM, and *Soft $F_\beta$* measure the fit of velocity value, subsurface structure, and geological interface, respectively. Thus results of our SeisInvNet ($\eta$) should have overall the best quality on these aspects (see Figs. 6 and 7 and the Supplementary Material for visual results).

In addition, we show the loss curve of baseline $\zeta$ and SeisInvNet ($\eta$) in Fig. 10. As can be seen, both methods converged quickly without overfitting on both training and validation sets, but our SeisInvNet ($\eta$) converged more rapidly and more stably, which means SeisInvNet ($\eta$) could cope with this task without too much burden and may well apply to other more complex data. It is worthwhile to mention again that our SeisInvNet achieves this promising performance with just one-quarter of the parameters used in the baseline model, which further proves our network's efficiency.

TABLE II
PERFORMANCE STATISTICS OF SEISINVNET ($\eta$) WITH AND WITHOUT GLOBAL CONTEXT, ON THE TEST SET, BY THE FIVE METRICS. FOR EACH METRIC, THE BETTER RESULTS ARE HIGHLIGHTED IN BOLD FONTS. THE ↑ INDICATES THE LARGER VALUE ACHIEVED, THE BETTER PERFORMANCE IS, WHILE ↓ INDICATES THE SMALLER, THE BETTER

| Dataset | Metric | SeisInvNet ($\eta$) | |
|---|---|---|---|
| | | without global | with global |
| Test | MAE ↓ | 0.019984 | **0.014962** |
| | MSE ↓ | 0.001720 | **0.001226** |
| | SSIM ↑ | 0.942725 | **0.953380** |
| | MSSIM ↑ | 0.981283 | **0.986100** |
| | Soft $F_\beta$ ↑ | 0.738418 | **0.761035** |

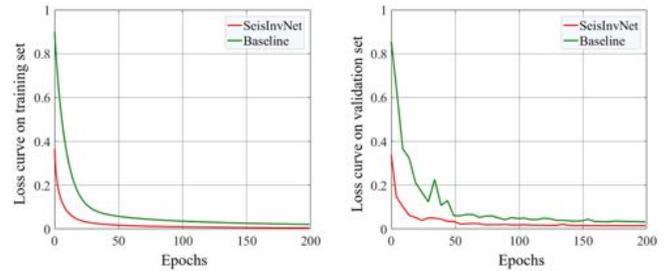

Fig. 10. Loss curves for baseline ($\zeta$) and SeisInvNet ($\eta$) on (Left) training set and (Right) validation set.

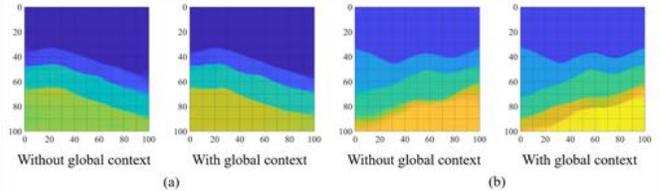

Fig. 11. SeisInvNet ($\eta$) outcomes with/without the global context. (a) and (b) correspond to the type III and IV model in Fig. 7.

Next, we would like to demonstrate is the effectiveness of the global context vector we added in the enhanced seismic trace. To check this, we retrain SeisInvNet ($\eta$) without the global context and show the statistic evaluation in Table II and visual examples in Fig. 11. From the experiments, without the global context, the performance degenerates significantly. This means that the global context contains some information that local seismic trace does not have and could benefit the reconstruction of the velocity model. Since the global context is extracted from the whole seismic profile, we guess it contains information such as velocity distribution and number of interfaces. However, what kind of information the global context actually has is hard to verify, and explanation of features in deep learning is still an open question.

Finally, in Fig. 12, we demonstrate how the hyperparameters applied affect the final results. The main hyperparameters used for our Adam optimizer [58] are the learning rate and batch size. From the loss curves in Fig. 12, our SeisInvNet ($\eta$) demonstrates certain robustness to these parameters.

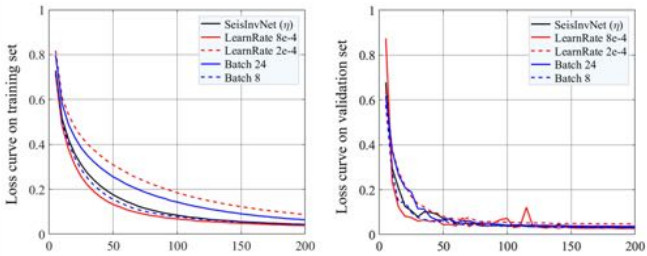

Fig. 12. Loss curves for SeisInvNet ($\eta$) with different hyperparameter settings on (left) training set and (right) validation set.

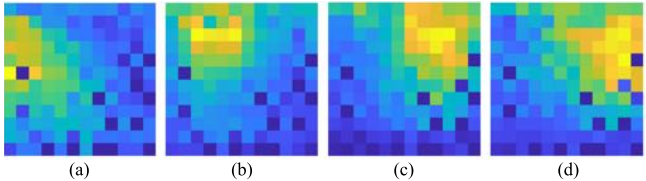

Fig. 13. Feature map visualization. We visualize feature maps that follow the approach described in Section IV-F, and each feature map is amplified for better illustration. Feature maps show clear sensitivity regions (the yellow areas). (a) 1–8 receivers. (b) 9–16 receivers. (c) 17–24 receivers. (d) 25–32 receivers.

### F. Mechanism Study

As stated in the previous sections, the benefit of extracting spatially aligned feature map from enhanced seismic traces lies in taking full advantage of knowledge inherent in each seismic trace. Typically, each enhanced seismic trace will have different sensitivity regions on the velocity model, so a feature map from a different enhanced seismic trace would reflect knowledge focus on the different parts of the velocity model. For simplicity, we also call the focused part in a feature map the sensitivity region. In Fig. 13, we illustrate how spatially aligned features $\mathbf{F}^i$ of SeisInvNet ($\eta$) look like. Since $D^i$ is of size [20, 1000, 32], we will have 640 recorded seismic traces and thus generate 640 feature maps. To save space, we show only the 32 feature maps for the 11th seismic source (the middle seismic source). Moreover, to visualize from an overall perspective, for each feature map, we compute its average over all the 1000 validation data, then group and further average every eight feature maps to better explore the pattern.

It can be easily seen from Fig. 13 that the feature map from recorded data by left receivers (#1–8) has high activation on the left-hand side, which means that it can provide more information to reconstruct left part of the velocity model. Similarly, feature maps by middle-left receivers (#9–16), middle-right receivers (#17–24), and right receivers (#25–32) own information to reconstruct middle-left, middle-right, and right part of the velocity model, respectively. However, there is little response to the bottom part of the velocity model in both feature maps, which means it is hard to infer information from seismic data to reconstruct bottom part and implies that the sensitivity region of feature maps cannot cover the whole velocity model.

Yet, most results of our SeisInvNet ($\eta$) look as good as Fig. 7, so we suspect that our inversion results for the bottom part of the velocity model may depend more on the context and the statistics information of data set that SeisInvNet had captured and memorized. The utilization of context and statistics is the primary advantage of DNN-based methods, but in the task of geophysical inversion, inversion for the bottom part in the velocity model by DNNs maybe still unstable and has bigger error due to the lack of direct knowledge. Basically, inversion of the bottom part is an inherent difficulty in the seismic inversion due to seismic inversion mechanism.

We study and verify this potential problem via Fig. 14, which illustrates the velocity difference and variance in different layers of the velocity model in the validation set, for the seven methods described in Section IV-C. The velocity difference is defined as the mean estimated velocity minus the mean true velocity within the area in each layer of the true velocity model. The velocity variance measures the variance of estimated velocity within the area in each layer of the true velocity model. From the velocity difference in Fig. 14, we can conclude that all the seven methods tend to generate higher velocity values for shallow layers and lower velocity values for deeper layers. As for velocity variance, the bars corresponding to the top and bottom layers have lower height than the bars corresponding to the middle layers, which indicates a stable and uniform velocity distribution within the top and bottom layers. In general, variants of SeisInvNet present results with smaller velocity difference and variance within each layer, so that they could give more accurate, stable, and uniform seismic inversion results. Similar observation can also be made in the test set.

### G. Network Generalization

Learning-based methods always have the problem of generalization. Directly applying the trained networks from one data set to another new data set usually will not lead to satisfying results. However, the trained networks often have gained some general knowledge and could be applied to another data set well after fine tuning or other operations. Generally, fewer the efforts made for these operations, much stronger the generalization ability of the trained networks is. In the following, we further carry out two experiments to demonstrate the generalization ability of our SeisInvNet ($\eta$). These two experiments will confirm that our SeisInvNet ($\eta$) has the potential of dealing with more complex models and noisy data. Moreover, it may provide a novel perspective for the data processing of incomplete seismic data that are often encountered in field cases. More visual results can be found in the Supplementary Material.

*1) Domain Adaptation:* In this experiment, we intend to adapt our SeisInvNet ($\eta$) to another data domain with two kinds of data that have never been seen before (noisy data of models with fault and noisy data of models with more layers). In this data set [SeisInv (new)], we have 2000 data-model pairs for each kind, which is only one-fifth of our previous SeisInv (old) data set. The average SNR for all the noisy data equals $-0.0716$ dB (we use Gaussian noise with zero-mean). To make domain adaptation from SeisInv (old) to SeisInv (new), we fine-tune SeisInvNet ($\eta$) by retraining it on SeisInv (new) for another 40 epochs (only one-fifth of initial training). The results are shown in Fig. 15. The first two rows illustrate how the results on noisy data of models with fault and models with more layers get promoted after fine-tuning, respectively. The third row is the result of retrained networks on the previously used model as shown in Fig. 7, from which most of the knowledge obtained from the initial training process had not been disregarded after fine-tuning, though there is a slight degradation in the last layer. Furthermore, we show the statistical evaluation in Table III in which the performance of SeisInvNet ($\eta$) before and after fine-tuning, on the new and old data sets, by the five metrics is demonstrated. Clearly, after fine-tuning, the performance of SeisInvNet ($\eta$) on SeisInv (new) is promoted significantly, while its degradation on SeisInv (old) is relatively small. This minor degradation problem could be further alleviated by joint training on the two data sets; however, here, we focus on domain adaptation.

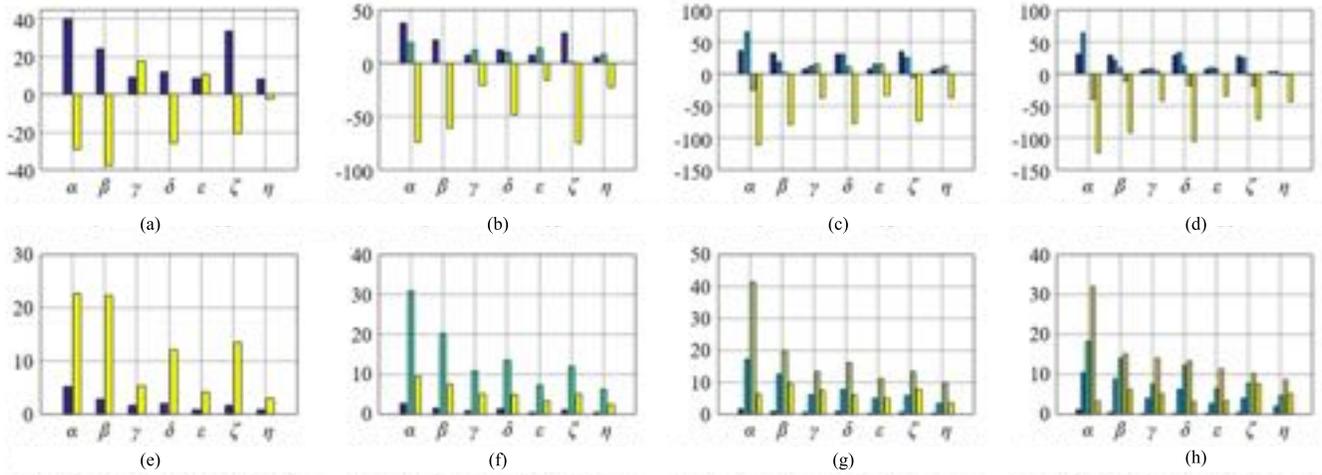

Fig. 14. Velocity difference (a)–(d) and velocity variance (e)–(h). From left to right, we show the results for the velocity model of four types I–IV, respectively. In each subfigure, we have seven groups of bars for seven methods ($\alpha$) to ($\eta$), respectively. For each group, bars from left to right correspond to the layers from top to bottom.

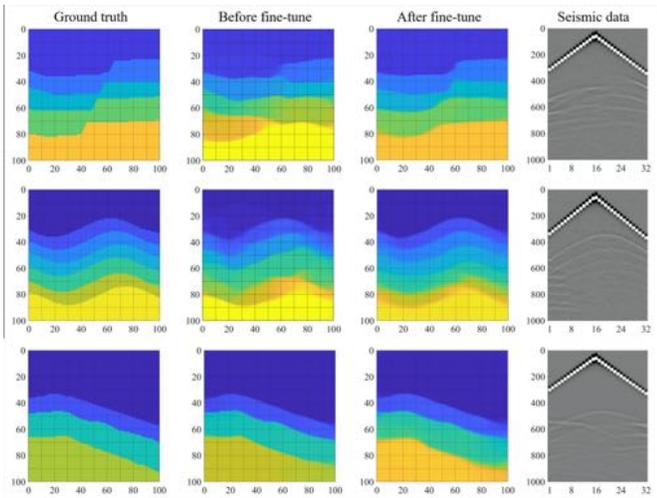

Fig. 15. Inversion results before and after fine-tuning. The top two rows are examples from SeisInv (new), whereas the last one is from SeisInv (old). The corresponding data are in the last column, where the first two are the noisy data with SNR equals $-0.0108$ and $-0.0603$ dB, respectively.

TABLE III
PERFORMANCE STATISTICS OF SEISINVNET ($\eta$) BEFORE AND AFTER FINE TUNING, ON THE NEW AND OLD DATA SETS, BY THE FIVE METRICS. FOR EACH SET AND METRIC, THE BETTER RESULTS ARE HIGHLIGHTED IN BOLD FONTS. THE ↑ INDICATES THE LARGER VALUE ACHIEVED, THE BETTER PERFORMANCE IS, WHILE ↓ INDICATES THE SMALLER, THE BETTER

| Dataset | Metric | SeisInvNet ($\eta$) | |
|---|---|---|---|
| | | before fine-tune | after fine-tune |
| SeisInv (new) | MAE ↓ | 0.087181 | **0.032341** |
| | MSE ↓ | 0.016998 | **0.003235** |
| | SSIM ↑ | 0.676609 | **0.863192** |
| | MSSIM ↑ | 0.762080 | **0.945074** |
| | Soft $F_\beta$ ↑ | 0.265253 | **0.582547** |
| SeisInv (old) | MAE ↓ | **0.014962** | 0.027404 |
| | MSE ↓ | **0.001226** | 0.002407 |
| | SSIM ↑ | **0.953380** | 0.924968 |
| | MSSIM ↑ | **0.986100** | 0.971589 |
| | Soft $F_\beta$ ↑ | **0.761035** | 0.701299 |

From Table III, we can conclude that our SeisInvNet ($\eta$) could apply to other data domain well after fine-tuning with much fewer training data and training epochs. Also, most of the knowledge SeisInvNet ($\eta$) learned from original SeisInv data set could still be preserved. These examples suggest that the trained SeisInvNet ($\eta$) has promising generalization ability and could apply to another data set with fewer efforts.

*2) Observation Setup:* In addition to the data domain, another generalization problem worth discussing is trained network's applicability on different observation setups. Actually, by the logic of reconstructing model from each enhanced seismic trace, our SeisInvNet is relative flexible to the observation setup. Especially, in our training stage, we had randomly dropped out 20% feature maps from which we decoded the velocity model (see Fig. 4). So, typically, our SeisInvNet could still work well without some seismic traces. To verify this claim, in the following, we carry out another experiment by randomly dropping out some receivers.

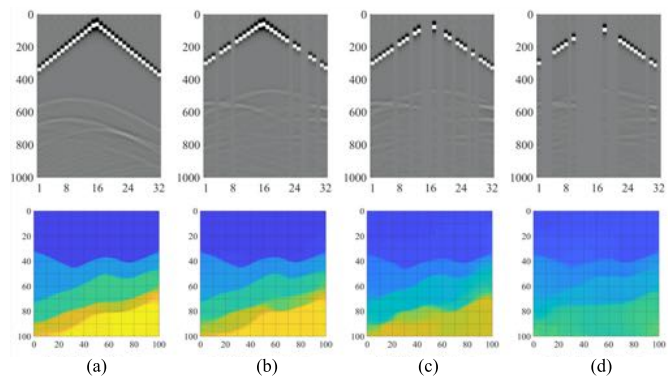

Fig. 16. Inversion results on different observation setups. Different numbers of receivers are randomly selected from the original 32 receivers. (a) 32 receivers. (b) 26 receivers. (c) 22 receivers. (d) 16 receivers.

As shown in Fig. 16, our SeisInvNet ($\eta$) could still present acceptable results when six seismic traces missed. However, the interfaces start to become blurred and misaligned after dropping out ten seismic traces. When half of the seismic traces are discarded, our SeisInvNet ($\eta$) could still perceive interface location and velocity distribution but with low accuracy. The results are acceptable and predictable, since we had only let networks to recover the model when 20% feature maps are missing.

In a word, the overall analysis proves a certain generalization ability of our SeisInvNet on the more complex data set and different observation setups. However, the generalization of deep-learning-based inversion methods on real data is still a challenging job and considering the physics may be one potential solution that we intend to study in our further works.

## V. CONCLUSION

In this article, we investigated the problem of deep-learning-based seismic inversion. We found that most of the existing approaches could not take full advantage of seismic data and may introduce biases. In light of these drawbacks, we analyze the intrinsic characteristics of this task and propose an end-to-end DNNs called SeisInvNet. Our SeisInvNet generates spatially aligned features from every enhanced seismic trace, which enforce every seismic trace to contribute to the reconstruction of the whole velocity model. In our experiments, these spatially aligned features actually show spatial correspondence to the velocity model, thus paving the way for subsequent components to regress velocity value by CNNs directly.

We apply $L_2$ and MSSIM loss together to guide the SeisInvNet reconstruct velocity model considering individual position and local structures together so as to generate results closer to the target. We propose to do evaluation from the aspects of the velocity value, subsurface structure, and geological interface and consequently choose five relevant metrics. Among these metrics, the proposed *Soft $F_\beta$* is a relaxed version of *F-measure* that can better measure the alignment of interfaces. To train, validate, and test our SeisInvNet, we synthesize and collect SeisInv data set with 12 000 pairs of seismic data and velocity model.

Our experiments show that our SeisInvNet demonstrates superior performance against all the baseline models on SeisInv data set by five metrics as well as visual comparison, consistently. With fewer parameters, our SeisInvNet achieves lower loss and converges more quickly and stably. In the mechanism study, we further provide the evidence to verify our statements. Finally, the study of network generalization proves certain application prospects of our SeisInvNet on more complex data, which may be encountered in field cases.

## ACKNOWLEDGMENT

The authors thank the editor and referees for their constructive suggestions.

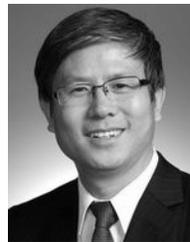

**Shucai Li** received the Ph.D. degree in rock and soil mechanics from the Institute of Rock and Soil Mechanics, Chinese Academy of Sciences, Beijing, China, in 1996.

After his graduation, he stayed and joined the institute as a researcher. Since 2001, he has been a Professor with the School of Civil Engineering and the Head of the Geotechnical and Structural Engineering Research Center, Shandong University, Jinan, China, where he is also with the School of Qilu Transportation. His research area is geophysical forward prospecting of adverse geology during tunneling.

Dr. Li serves as the Chief Editor for *Tunnelling and Underground Space Technology*.

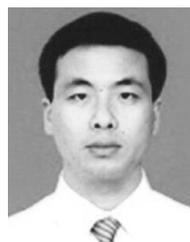

**Bin Liu** received the B.S. and Ph.D. degrees in civil engineering from Shandong University, Jinan, China, in 2005 and 2010, respectively.

He then joined the Geotechnical and Structural Engineering Research Center, Shandong University, where he is currently a Professor with the School of Qilu Transportation. His research area is engineering geophysical prospecting techniques, especially their applications in tunnels.

Dr. Liu is a member of the Society of Exploration Geophysicists (SEG) and the International Society for Rock Mechanics and Rock Engineering (ISRM), and serves as a Council Member for the Chinese Geophysical Society.

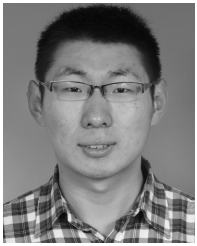

**Yuxiao Ren** received the bachelor's degree from Shandong University, Jinan, China, in 2014, and the master's degree in mathematics from Loughborough University, Loughborough, U.K., in 2015. He is currently pursuing the Ph.D. degree in civil engineering with Shandong University.

He is also a Visiting Scholar with the Georgia Institute of Technology, Atlanta, GA, USA, under the supervision of Prof. F. Herrmann. His research interests include seismic modeling and imaging, full-waveform inversion, and deep-learning-based geophysical inversion.

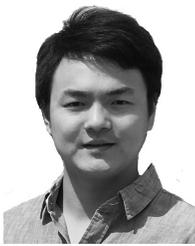

**Yangkang Chen** received the B.S. degree in exploration geophysics from the China University of Petroleum, Beijing, China, in 2012, and the Ph.D. degree in geophysics from The University of Texas at Austin, Austin, TX, USA, in 2015, under the supervision of Prof. S. Fomel.

He then joined the Oak Ridge National Laboratory, Oak Ridge, TN, USA, as a Distinguished Post-Doctoral Research Associate and conducted research on global adjoint tomography. He is currently an Assistant Professor with Zhejiang University, Hangzhou, China. He has published more than 100 internationally renowned journal articles and more than 80 international conference papers. His research interests include seismic signal analysis, seismic modeling and inversion, simultaneous-source data deblending and imaging, global adjoint tomography, and machine learning.

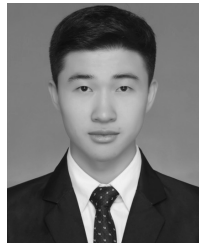

**Senlin Yang** received the bachelor's degree in engineering from Shandong University, Jinan, China, in 2017, where he is currently pursuing the Ph.D. degree.

His research interest includes deep-learning-based geophysical data processing and inversion.

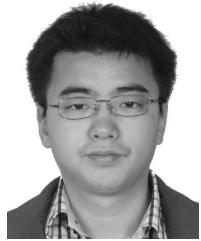

**Yunhai Wang** (M'17) received the Ph.D. degree in computer science from the Supercomputer Center, Chinese Academy of Sciences, Beijing, China, in 2011.

He is currently a Professor with the School of Computer Science and Technology, Shandong University, Jinan, China. His research interests include scientific visualization, information visualization, and computer graphics.

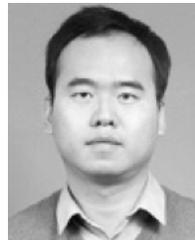

**Peng Jiang** (S'12–M'18) received the B.S. and Ph.D. degrees in computer science and technology from Shandong University, Jinan, China, in 2010 and 2016, respectively.

He is currently a Research Assistant with the School of Qilu Transportation, Shandong University. He has published many works on top-tier venues, including the International Conference on Computer Vision, the Conference on Neural Information Processing Systems, and the IEEE TRANSACTIONS ON IMAGE PROCESSING. Recently, he is focusing on deep-learning-based geophysical inversion. His research spans various areas, including computer vision, image processing, machine learning, and deep learning.